\documentclass[letterpaper, 10 pt, conference]{ieeeconf}  

\IEEEoverridecommandlockouts                              

\usepackage{graphics} 
\usepackage{epsfig} 
\usepackage{mathptmx} 
\usepackage{times} 
\usepackage{amsmath} 
\usepackage{amssymb}  
\usepackage{multirow}
\usepackage{booktabs}
\usepackage{ulem} 
\usepackage{cite}
\usepackage{stfloats}
\usepackage{xcolor}
\usepackage{makecell}
\usepackage[ruled,vlined,linesnumbered]{algorithm2e} 

\usepackage[caption=false]{subfig}
\usepackage{hyperref}
\hypersetup{hidelinks,
	colorlinks=true,
	allcolors=black,
	pdfstartview=Fit,
	breaklinks=true}
\newcommand{\warning}[1]{\textcolor[RGB]{0, 0, 0}{#1}}

\title{\LARGE \bf
SFCo-Nav: Efficient Zero-Shot Visual Language Navigation via Collaboration of Slow LLM and Fast Attributed Graph Alignment 
}

\author{
Anonymous Authors
}

\author{
Chaoran Xiong$^{1,\dagger}$,~\IEEEmembership{Graduate~Student~Member,~IEEE}, Litao Wei$^{1,2,\dagger}$, Xinhao Hu$^{1,\dagger}$, Kehui Ma$^1$,\\ Ziyi Xia$^1$, Zixin Jiang$^{1}$, Zhen Sun$^1$, and Ling Pei$^{1,3,\ast}$,~\IEEEmembership{Senior~Member,~IEEE}
\thanks{$^{\dagger}$Chaoran Xiong, Litao Wei, and Xinhao Hu contribute equally to this work.}
\thanks{$^{\ast}$Corresponding Author: Ling Pei.}
\thanks{This work was supported in part by the National Natural Science Foundation of China (NSFC) under Grant No. 62273229, and in part by the Science and Technology Commission of Shanghai Municipality under Grant Nos. 24DZ3101300, 24TS1402600, and 24TS1402800.}
\thanks{$^1$The authors are with the Shanghai Key Laboratory of Navigation and Location Based Services, Shanghai Jiao Tong University (SJTU), Shanghai 200240, China. $^2$Litao Wei is also with Zhiyuan College, SJTU. $^3$Ling Pei is also with the State Key Laboratory of Submarine Geoscience, SJTU. (e-mail: \{sjtu4742986; oscar0731; xinhaohu; khma0929; matcha.latte; zhensun; yan.xiang; ling.pei\}@sjtu.edu.cn; jiangzixin0214@163.com).}
\thanks{Demo and code will be released at \url{https://anonymous.4open.science/r/GQ-Nav-5EB5/}.}
}

\begin{document}

\maketitle
\thispagestyle{empty}
\pagestyle{empty}

\begin{abstract}
Recent advances in large vision-language models (VLMs) and large language models (LLMs) have enabled zero-shot approaches to visual language navigation (VLN), where an agent follows natural language instructions using only ego perception and reasoning. However, existing zero‑shot methods typically construct a naive observation graph and perform per‑step VLM–LLM inference on it, resulting in high latency and computation costs that limit real‑time deployment. To address this, we present SFCo-Nav, an efficient zero-shot VLN framework inspired by the principle of slow–fast cognitive collaboration. SFCo-Nav integrates three key modules: 1) a slow LLM-based planner that produces a strategic chain of subgoals, each linked to an imagined object graph; 2) a fast reactive navigator for real-time object graph construction and subgoal execution; and 3) a lightweight asynchronous slow–fast bridge aligns advanced structured, attributed imagined and perceived graphs to estimate navigation confidence, triggering the slow LLM planner only when necessary. To the best of our knowledge, SFCo-Nav is the first slow-fast collaboration zero-shot VLN system supporting asynchronous LLM triggering according to the internal confidence. Evaluated on the public R2R and REVERIE benchmarks, SFCo‑Nav matches or exceeds prior state‑of‑the‑art zero‑shot VLN success rates while cutting total token consumption per trajectory by over 50\% and running more than 3.5$\times$ faster. Finally, we demonstrate SFCo‑Nav on a legged robot in a hotel suite, showcasing its efficiency and practicality in indoor environments.
\end{abstract}

\section{INTRODUCTION}
Visual language navigation (VLN) requires an embodied agent to follow natural-language navigation instructions by perceiving the environment, reasoning about the instructions, and executing a sequence of actions to reach a goal\cite{R2R}. It is a fundamental task of a general embodied navigation (EN) system\cite{EmbodiedNav}. 

Recently, the success of large-scale Vision-Language Models (VLMs)\cite{gpt4V} and Large Language Models (LLMs)\cite{DSV3,GPT4} has inspired the development of zero-shot VLN systems\cite{opennav, vlmaps, NavGPT, MapGPT, CityNav}. These zero-shot methods exploit the pretrained reasoning and grounding capabilities of these models without task-specific fine-tuning. In contrast to conventional end-to-end trained VLN agents\cite{navila,DUET,HAMT}, zero-shot VLN offers significant advantages, such as no training cost, rapid deployment in new environments, and strong generalization to unseen instructions and layouts\cite{navcot}.

The core challenge in zero-shot VLN lies in efficiently transforming visual observations and textual instructions into accurate navigation actions. Early work, such as NavGPT\cite{NavGPT}, preprocesses all navigable viewpoints with a VLM BLIP-2\cite{Blip-2} to produce textual descriptions of the scene. Then the agent queries an LLM to output the next target viewpoint ID. While NavGPT verified LLMs’ navigation capability, it is an offline system that demands heavy computation. More recently, MapGPT\cite{MapGPT} employs a naive structured topological map representation and integrated VLMs like GPT-4V\cite{gpt4V} or BLIP-2\cite{Blip-2} for online, step-by-step image-to-text conversion and direct action prediction. Although online-capable, this approach still requires per-step VLM processing over multiple candidate viewpoints. This leads to substantial token usage and slow inference. To achieve higher efficiency, NavCoT\cite{navcot} partially mitigates the cost by fine-tuning the LLM to reduce token consumption, yet retains the same expensive VLM multi-point perception at each step.

\begin{figure}
    \centering
    \includegraphics[width=1.0\linewidth]{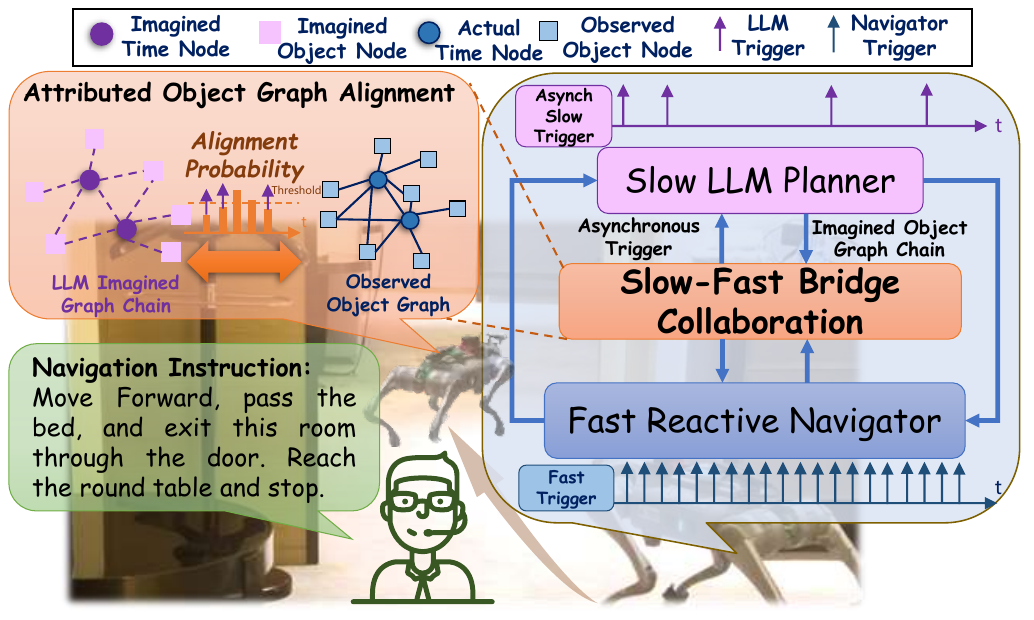}
    \caption{SFCo-Nav is an efficient zero-shot VLN framework inspired by slow–fast cognitive collaboration. It comprises three modules: 1) a slow brain LLM-based planner; 2) a fast brain reactive navigator; and 3) a lightweight asynchronous slow–fast bridge that aligns the imagined and perceived graphs, estimates navigation confidence, and triggers LLM only when necessary. This design minimizes costly LLM calls while preserving high navigation success.}
    \label{fig:abstract}
    \vspace{-0.6cm}
\end{figure}

Current zero-shot VLN methods\cite{NavGPT, MapGPT, CityNav, navcot, vlmaps, opennav} usually adopt VLM–LLM paradigm. This paradigm can be referred to as a full slow-brain strategy: at every navigation step, the system exhaustively queries a VLM for all visible viewpoints and uses an LLM to directly decide the next action. This results in high token cost and slow decision process. Conversely, human navigators typically perform an initial phase of slow thinking: parsing instructions, setting subgoals, and forming an internal plan. Then the fast thinking executes the subgoals using perceptual intuition. Slow reasoning is invoked only when confidence drops in unfamiliar or altered situations. Details may refer to \cite{ThinkingFastSlow}. Such a slow–fast collaboration mechanism has the potential to improve navigation efficiency, but has been overlooked in zero-shot VLN.

Inspired by human slow–fast cognitive synergy\cite{ThinkingFastSlow}, we present SFCo‑Nav, a zero‑shot VLN framework combining an LLM‑based slow brain with a lightweight reactive fast brain for efficient embodied navigation. The slow brain parses instructions into a strategic decision chain including target objects, navigation skills, and an imagined object graph. On the other hand, the fast brain perceives a real‑time object graph and executes the plan using object‑skill primitives. An asynchronous triggering mechanism monitors a structured attributed graph\cite{AGA} matching confidence and engages the slow brain only when confidence drops below a threshold, minimizing unnecessary reasoning. SFCo‑Nav achieves comparable or higher success rates than strong zero‑shot baselines on R2R\cite{R2R} and REVERIE\cite{Reverie}, with significantly lower inference costs. Additionally, SFCo‑Nav demonstrates practical efficiency in a real‑world hotel‑suite deployment on a legged robot. To the best of our knowledge, it is the first slow‑fast zero‑shot VLN system with asynchronous LLM triggering based on internal confidence.  
\warning{Our main contributions are as follows}:
\begin{enumerate}
    \item 
    A zero-shot slow-fast collaborative navigation framework SFCo-Nav for embodied navigation. Our slow component, an LLM, generates a strategic decision chain of target objects, required skills and imagined object graph. The fast component, a reactive navigator, then executes these steps by efficiently aligning a real-time, perceived object graph with the target objects and imagined object graphs generated by LLM.
    \item 
    A lightweight, asynchronous triggering mechanism that governs the slow-fast collaboration for superior efficiency. This is achieved by computing an advanced structured, attributed graph matching probability as the navigator confidence based on the alignment of the perceived graph with the LLM imagined graph chain. 
    \item
    State-of-the-art efficiency in zero-shot embodied navigation, demonstrated on the R2R and REVERIE benchmarks. SFCo-Nav achieves comparable or higher task success rates than existing methods while reducing average total consumed tokens per trajectory by over 50\% and running more than 3.5$\times$ faster.
\end{enumerate}

The remainder of this paper is organized as follows. Section II reviews related work on zero-shot VLN and slow–fast systems. Section III formalizes the proposed slow–fast collaboration framework for VLN. Section IV presents the architecture and technical details of SFCo-Nav. Section V reports experimental results on public VLN benchmarks, including comparisons with state-of-the-art zero-shot methods, ablation analyses, and a real-world case study. Finally, a conclusion is given in Section VI.

\section{RELATED WORK}
In this section, we review relevant literature in zero‑shot VLN. Section II‑A examines recent LLM‑based approaches that leverage large‑scale pretrained models to interpret navigation instructions. Section II‑B discusses slow–fast system principles in perception and decision‑making, and their potential to improve the efficiency of zero‑shot VLN.  
\subsection{LLM-Based Zero-Shot VLN}
Advances in foundation Vision–Language Models (VLMs)\cite{gpt4V} and Large Language Models (LLMs)\cite{DSV3,GPT4} have enabled agents to follow natural language navigation instructions without task-specific training. Unlike conventional end-to-end VLN\cite{DUET, HAMT, Scalevln}, zero-shot VLN leverages pretrained models directly, avoiding costly data collection and fine-tuning, which is a key advantage for real-world deployment\cite{NavGPT, MapGPT, CityNav}.  

NavGPT\cite{NavGPT} first demonstrated LLM navigation capabilities by converting all viewpoints into BLIP-2 textual descriptions offline and prompting the LLM to select the next viewpoint. Despite its effectiveness, this per-step multi-view processing made online use impractical. A2Nav introduced Action Awareness so the LLM could reason over both scene observations and available navigation skills\cite{A2Nav}, while Console\cite{Console} used the LLM to filter and refine landmarks, boosting success but still relying on scene-trained VLN models.  

To enable online execution, MapGPT\cite{MapGPT} used a simple topological map linking navigable viewpoints, converting multiple observations per step into text for LLM-driven action prediction. However, the naive structure of scene description and multi-view VLM processing per step caused high token cost and latency. NavCoT\cite{navcot} reduced LLM token usage and improved reasoning via LLM fine-tuning, but retained the same expensive multi-view VLM computation, undermining the zero-shot advantage. These limitations highlight the need for a real-time, low-cost, deployable zero-shot VLN approach.

\subsection{Slow–Fast Systems}
Current zero-shot VLN methods, such as \cite{NavGPT, MapGPT, CityNav, navcot}, usually adopt a full slow-brain strategy, invoking VLM–LLM reasoning at every step to output actions. While comprehensive, this leads to redundant perception, high computation cost, and low time efficiency. Inspired by human fast and slow thinking\cite{ThinkingFastSlow}, slow–fast systems have gained attention in robotics, particularly in manipulation, where a slow high-level planner guides a fast low-level controller for efficient execution\cite{RDP}.  

In zero-shot visual language navigation scenario, however, slow–fast collaboration is largely unexplored. The longer distances and larger spatial scope demand efficient confidence-based triggers for re-planning. Existing methods\cite{NavGPT, MapGPT, MC-GPT} either invoke slow reasoning at every step or lack adaptive mechanisms, both of which incur high latency. This gap motivates our design of a slow–fast collaborative zero-shot VLN framework that reduces slow LLM reasoning overhead for practical deployment.

\begin{figure*}[t]
    \centering
    \includegraphics[width=0.8\linewidth]{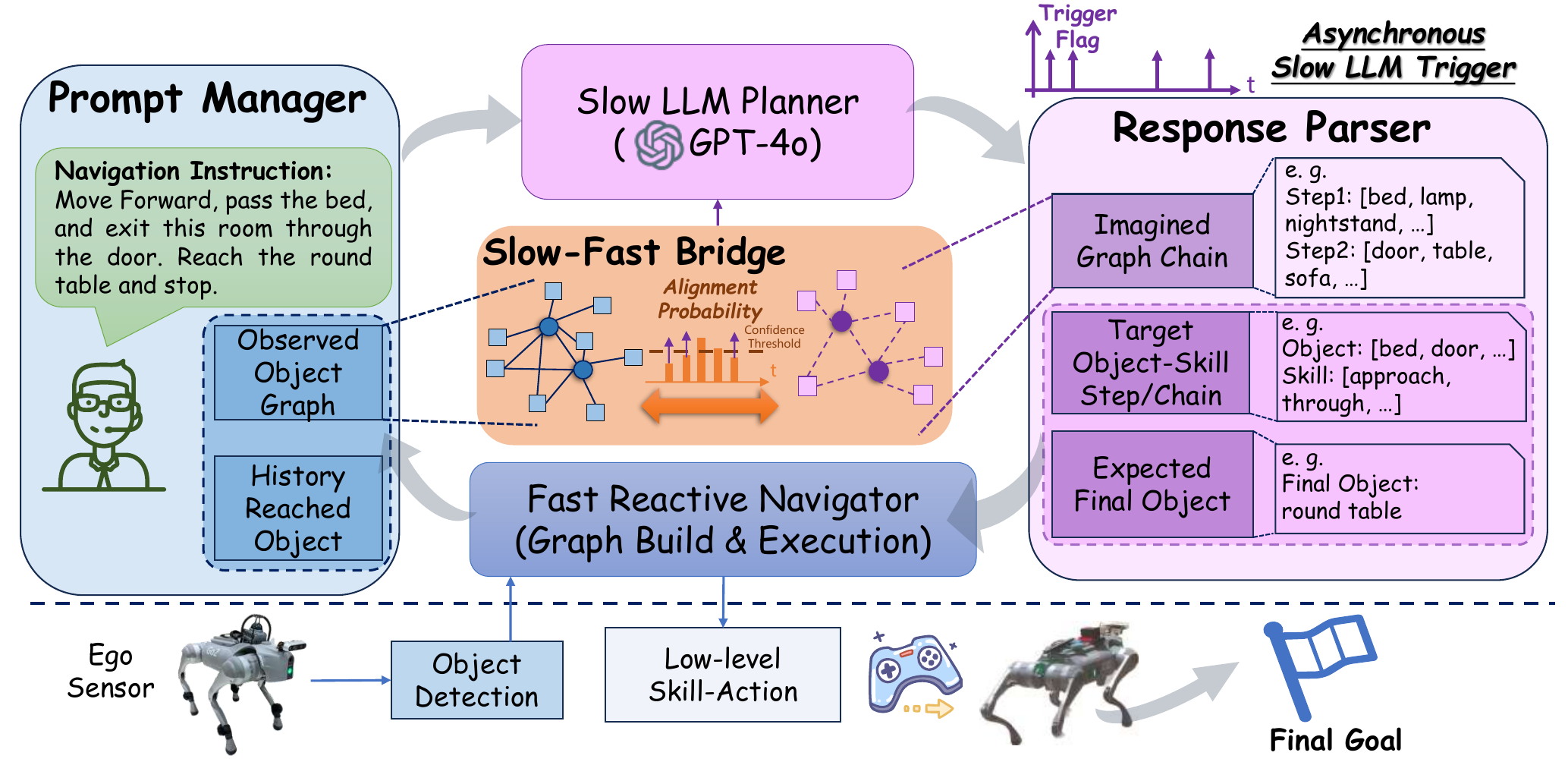}
    \caption{
    System overview of SFCo‑Nav, a slow–fast collaborative framework for efficient zero-shot visual language navigation. 
    The Slow LLM Planner (\(\Pi_{\mathrm{slow}}\)) decomposes the navigation instruction into subgoals, each paired with an imagined object graph \(G^{i}_t\). 
    The Fast Reactive Navigator (\(\pi_{\mathrm{fast}}\)) builds a perceived object graph \(G^{p}_t\) in real time and executes low-level actions to align with \(G^{i}_t\). 
    The Slow–Fast Bridge computes the graph-alignment confidence \(C_t\); high confidence (\(C_t > \tau_C\)) keeps control with \(\pi_{\mathrm{fast}}\), while low confidence (\(C_t \leq \tau_C\)) triggers replanning by \(\Pi_{\mathrm{slow}}\).
    }
    \label{fig:overview}
    \vspace{-0.5cm}
\end{figure*}
\section{PROBLEM FORMULATION}
\label{sec:problem_formulation}
To solve the slow-fast problem in VLN, our zero-shot slow–fast collaboration framework is formalized as follows.  
We consider a zero-shot navigation setting in which an embodied agent follows a natural language instruction \(I\) to reach a target object or location in an unknown environment.  
The task is to produce a sequence of low-level actions $\mathbf{A} = (a_0, a_1, \dots, a_T)$ 
that guides the agent to the instruction-specified goal while minimizing use of computationally expensive reasoning components.

At each timestep \(t\), the agent perceives the environment from egocentric sensory input and maintains an internal state \(S_t\) that summarizes observations and task progress.  
The navigation policy is realized as a hybrid slow–fast architecture as follows:
\begin{enumerate}
    \item Slow High-Level Planner (\(\Pi_{\mathrm{slow}}\)): 
    A high-capacity but costly reasoning module that, when invoked, generates or updates 
    a high-level plan \(\mathcal{P}_t\) from \(I\) and \(S_t\). 

    \item Fast Low-Level Controller (\(\pi_{\mathrm{fast}}\)): 
    A low-latency policy that executes local control actions using recent observations 
    and the current high-level plan \(\mathcal{P}_t\).

    \item Collaboration Module: 
    A decision mechanism that determines when to switch navigation policy between \(\pi_{\mathrm{fast}}\) 
    and \(\Pi_{\mathrm{slow}}\) based on agent internal indicators, which
    seeks to maximize task success while minimizing reliance on \(\Pi_{\mathrm{slow}}\).
\end{enumerate}

Navigation terminates when the agent reaches the goal within a stopping distance \(d_{\mathrm{stop}}\).  
The design objective is to jointly optimize \(\Pi_{\mathrm{slow}}\), \(\pi_{\mathrm{fast}}\), and the slow-fast bridge rule to maximize task success while reducing the reliance on the slow planner for efficiency.

\section{METHODOLOGY}
In this section, in order to address the zero‑shot navigation problem formulated in Section~\ref{sec:problem_formulation}, we present SFCo‑Nav, our proposed slow–fast collaborative framework. Firstly, the overall architecture of SFCo‑Nav is introduced. Then a detailed description of its three core components is provided.

\subsection{System Overview}
\label{sec:overview}

The architecture of SFCo‑Nav, shown in Fig.~\ref{fig:overview}, is designed around a slow–fast collaboration principle to achieve efficient embodied navigation. 
The system is composed of three primary modules:

\begin{itemize}
    \item Slow LLM Planner (\(\Pi_{\mathrm{slow}}\)) serves as the "slow brain," decomposing a user's instruction \(I\) into a sequence of subgoals.  
    Each subgoal contains a target object and a required skill, together with an imagined object graph \(G^{i}_t\), representing the LLM’s semantic and spatial prior for the expected subgoal scene.

    \item Fast Reactive Navigator (\(\pi_{\mathrm{fast}}\)) acts as the "fast brain," operating in real time.  
    It builds a local perceived object graph \(G^{p}_t\) from ego-centric sensor inputs (e.g., RGB-D) and outputs low-level actions \(a_t\) to greedily maximize the alignment between \(G^{p}_t\) and the subgoal’s \(G^{i}_t\).

    \item Slow–Fast Bridge implements asynchronous collaboration by computing a graph-alignment confidence score \(C_t = \mathrm{Align}(G^{p}_{1:t}, G^{i}_{1:t})\).  
    If \(C_t > \tau_C\), control remains with \(\pi_{\mathrm{fast}}\).  
    Otherwise, the bridge triggers \(\Pi_{\mathrm{slow}}\) to replan.
\end{itemize}

The following subsections detail the design and implementation of each of these three modules.

\subsection{Slow LLM Planner: Object-Skill-Graph Chain Generation}

\begin{figure}
    \centering
    \includegraphics[width=1.0 \linewidth]{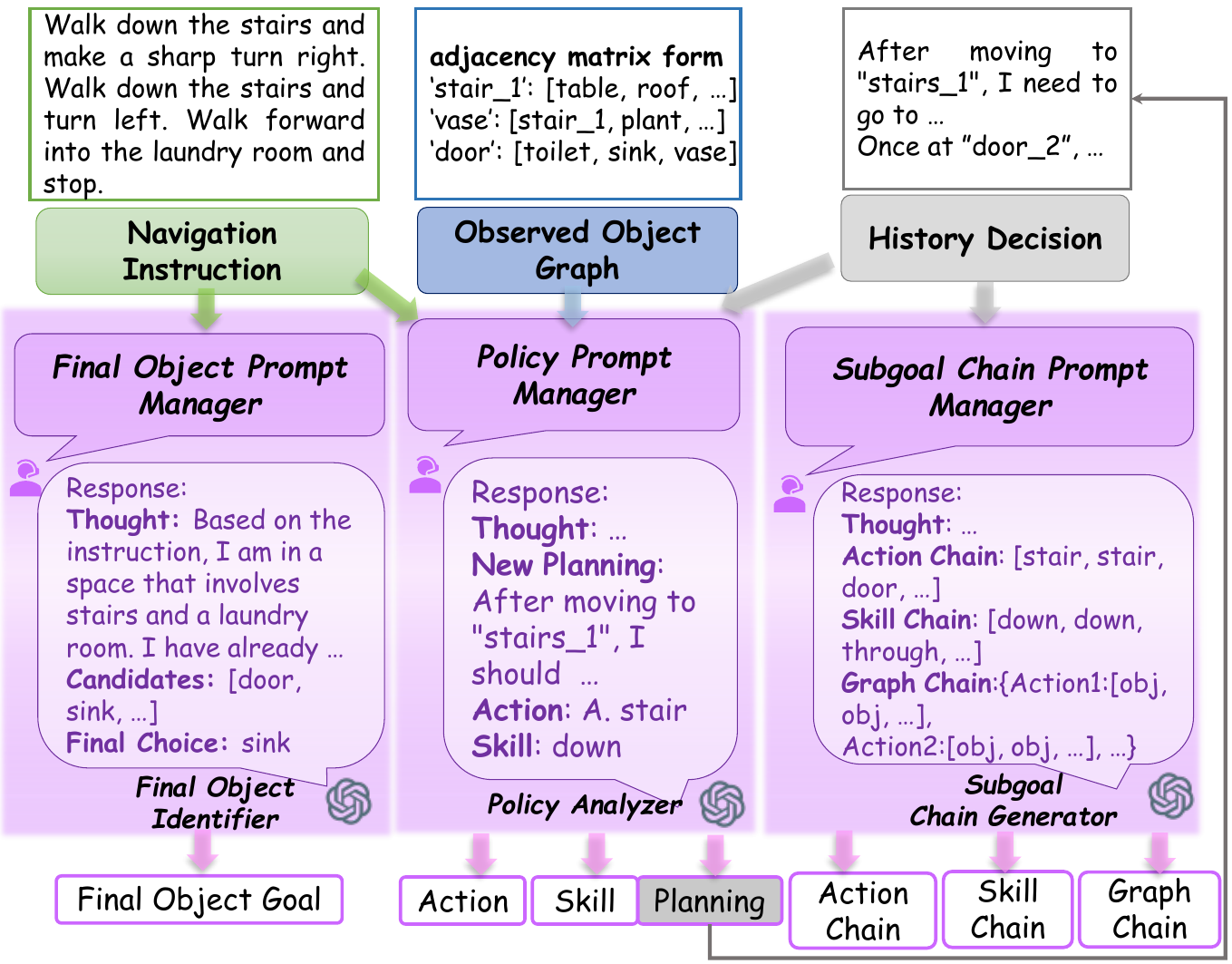}
    \caption{Slow LLM Planner prompt structure and operation process.}
    \label{fig:llm}
    \vspace{-0.5cm}
\end{figure}

Let the user's navigation instruction be denoted by a string \(I\). 
The agent's perception of the environment is represented by a perceived object graph \(G^{p}_t\). 
The full navigation plan generated by the Slow Planner \(\Pi_{\mathrm{slow}}\) is denoted by \(\mathcal{P}\).
As show in Fig. \ref{fig:llm} The Slow LLM Planner operates as a sequential pipeline of three functional modules:

\subsubsection{Final Object Identifier (\(f_{\mathrm{goal}}\))}
This initial module parses the user's command \(I\) to determine the final navigation target \(o_{\mathrm{goal}}\):
\begin{equation}
    o_{\mathrm{goal}} = f_{\mathrm{goal}}(I).
    \label{eq:f_goal}
\end{equation}
The LLM is prompted to extract the object name where the navigation task concludes. 
This output defines the agent's global stopping condition. The task is successful if the agent's state \(s_T\) at the final timestep \(T\) satisfies 
\(\mathrm{distance}(s_T, o_{\mathrm{goal}}) \leq d_{\mathrm{stop}}\), 
where \(d_{\mathrm{stop}}\) is a predefined threshold (e.g., 3\,m).

\subsubsection{Policy Analyzer (\(f_{\mathrm{policy}}\))}
This module generates the agent's immediate subgoal and long-horizon strategic plan based on the current context. 
Let \(H_{t-1} = \{o_1, o_2, \dots, o_{t-1}\}\) denote the history of previously reached subgoal objects. 
The module is conditioned on the instruction \(I\), history \(H_{t-1}\), and current perceived object graph \(G^{p}_t\):
\begin{equation}
    (R_t, (o_t, sk_t)) = f_{\mathrm{policy}}(I, H_{t-1}, G^{p}_t).
    \label{eq:f_policy}
\end{equation}
Here, \(o_t\) is the immediate target object and \(sk_t\) is the required skill to reach it. 
The reasoning trace \(R_t\) outlines the high-level plan in natural language, while the subgoal \((o_t, sk_t)\) is an actionable command for execution in the next phase.

\subsubsection{Subgoal Chain Generator (\(f_{\mathrm{chain}}\))}
\label{sec:chain_generator}
This module translates the human-readable reasoning trace \(R_t\) into a structured, machine-executable subgoal chain \(\mathcal{P}_t\):
\begin{equation}
    \mathcal{P}_t = f_{\mathrm{chain}}(R_t).
    \label{eq:f_chain}
\end{equation}
The plan \(\mathcal{P}_t\) consists of \(N\) future subgoals:
\begin{equation}
    \mathcal{P}_t = [ (o_{t+j}, sk_{t+j}, G^{i}_{t+j}) ]_{j=0}^{N-1},
    \label{eq:plan_structure}
\end{equation}
where \(o_{t+j}\) is the target object, \(sk_{t+j}\) is the required skill, and \(G^{i}_{t+j}\) is the LLM-generated imagined object graph, representing a structured semantic–spatial prior for the expected scene at subgoal \(o_{t+j}\). 
The first tuple \((o_t, sk_t, G^{i}_t)\) corresponds to the immediate subgoal to be executed by \(\pi_{\mathrm{fast}}\) unless the Slow–Fast Bridge triggers \(\Pi_{\mathrm{slow}}\) based on low confidence.

\subsection{Fast Reactive Navigator: Object Graph Construction and Skill Planner}

The fast reactive navigator is responsible for grounding the Slow Planner's (\(\Pi_{\mathrm{slow}}\)) abstract commands into low-level actions. It operates in a tight perception–action loop with two primary functions: 1) constructing a real-time representation of the environment as a perceived object graph, and 2) executing the current LLM-defined subgoal based on this representation.

\subsubsection{Perceived Object Graph Construction}
\begin{figure}
    \centering
    \includegraphics[width=0.85\linewidth]{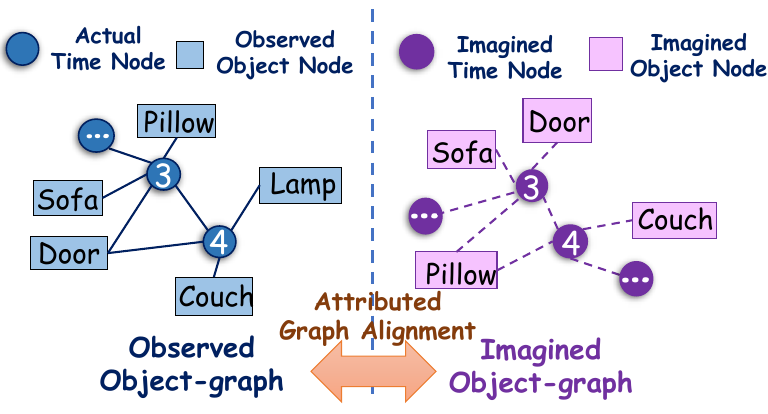}
    \caption{Observed and imagined attributed graph structure.}
    \label{fig:graph}
\end{figure}

At each timestep \(t\), the navigator builds a dynamic, attributed graph $G^{p}_t = (V_t, E_t)$, to represent its local understanding of the scene, as illustrated in Fig. \ref{fig:graph}. This graph includes the agent itself as a special reference node. The set of nodes \(V_t\) consists of a timestep node \(v_{\mathrm{timestep}}\) representing the agent, and a node \(v_j\) for each object detected in the current field of view. Each object node \(v_j\) is attributed with its semantic label (e.g., chair), its estimated 3D position relative to the agent, and the timestamp of its last observation. On the other hand, the edges \(E_t\) represent spatial relationships between nodes. For simplicity and efficiency, we construct a star-topology graph where edges connect the timestep node \(v_{\mathrm{timestep}}\) to each observed object node \(v_j\). Each edge \(e_{\mathrm{timestep}, j} \in E_t\) is attributed with the relative distance and bearing from the agent to the object.

Different from other zero-shot VLN methods using large-scale VLM for all navigable viewpoint description, our object graph is constructed by only one-time lightweight object detection for the current view of the agent, thus reducing token and time consumption in the visual perception process.

\subsubsection{Subgoal Execution Planner}
The planner’s goal is to execute the current subgoal \((o_t, sk_t, G^{i}_t)\) provided by the Slow Planner \(\Pi_{\mathrm{slow}}\). It implements a reactive policy \(\pi_{\mathrm{fast}}\) that maps the current perceived graph \(G^{p}_t\), the target object \(o_t\) and its corresponding skill \(sk_t\) to a low-level control action $a_t$. For instance, the subgoal execution planner can adopt skills such as approach, through or go up or go down to the target objects, which improves the navigator execution accuracy and efficiency.

\subsection{Slow-Fast Bridge: Attributed Object Graph Alignment}
The slow–fast bridge adopts a lightweight, asynchronous triggering mechanism that governs the slow–fast collaboration for superior efficiency. This is achieved by computing an advanced structured, attributed graph\cite{AGA} matching probability as the navigator confidence \(C_t\). This confidence score is based on the alignment of the perceived object graph \(G^{p}_t\) with the imagined object graph chain \(G^{i}_t\).

The vertices in the structured attributed graphs represent viewpoint timesteps as users \(\mathcal{V}_{\mathrm{u}}\) and objects as attributes \(\mathcal{V}_{\mathrm{a}}\).
To describe the probabilistic model for edges, we first consider the set of object–object vertex pairs \(\mathcal{E}_{\mathrm{u}} \triangleq \mathcal{V}_{\mathrm{u}} \times \mathcal{V}_{\mathrm{u}}\) and the set of object–viewpoint vertex pairs \(\mathcal{E}_{\mathrm{a}} \triangleq \mathcal{V}_{\mathrm{u}} \times \mathcal{V}_{\mathrm{a}}\). For any vertex pair \(e \in \mathcal{E} \triangleq \mathcal{E}_{\mathrm{u}} \cup \mathcal{E}_{\mathrm{a}}\), we write \(G^{p}_t(e)=1\) if an edge exists and \(G^{p}_t(e)=0\) otherwise (and similarly for \(G^{i}_t(e)\)). The joint observation of an edge pair is denoted \((G^{p}_t, G^{i}_t)(e)\).
\begin{algorithm}[t]
\label{alg:pq}
\SetKwInOut{Input}{Input}
\SetKwInOut{Output}{Output}
\Input{Perceived Graph $G_{1:t}^{p}$, Imagined Graph $G_{1:t}^{i}$}
\Output{Probability matrices $P$, $Q$}
\BlankLine
$\mathcal{T} \gets \text{time\_node}(G_{1:t}^{p}) \cup \text{time\_node}(G_{1:t}^{i})$\;

$\mathcal{O} \gets \text{objects}(G_{1:t}^{p}) \cup \text{objects}(G_{1:t}^{i})$\;
\BlankLine
$\begin{array}{cc}
Q = \begin{bmatrix} q_{00} & q_{01} \\ q_{10} & q_{11} \end{bmatrix} \gets \mathbf{0}_{2\times2} & 
P = \begin{bmatrix} p_{00} & p_{01} \\ p_{10} & p_{11} \end{bmatrix} \gets \mathbf{0}_{2\times2} \\
\end{array}$\;

\For{$(t,o) \in \mathcal{T} \times \mathcal{O}$}{
    $i \gets (t \in G_{1:t}^p \land o \in G_t^p)$\;
    $j \gets (t \in G_{1:t}^i \land o \in G_t^i)$\;
    $q_{ij} \gets q_{ij} + 1$\;
}

\For{$t_m,t_n \in \mathcal{T}, m<n$}{
     $i \gets \left(\exists o' \in \mathcal{O}: o' \in G_{t_m}^p \land o' \in G_{t_n}^p\right)$\;
    $j \gets \left(\exists o' \in \mathcal{O}: o' \in G_{t_m}^i \land o' \in G_{t_n}^i\right)$\;
   $p_{ij} \gets p_{ij} + 1$\;
}
$P \gets Norm_1(P),\; Q \gets Norm_1(Q) $ \;
\Return $P$, $Q$\;
\caption{$\boldsymbol{P},\boldsymbol{Q}$ Computation for Graph Alignment}
\end{algorithm}
The edges are generated according to a correlated model. For each object–object pair \(e \in \mathcal{E}_{\mathrm{u}}\), the edge probabilities \(\boldsymbol{P} = (p_{11}, p_{10}; p_{01}, p_{00})\) are given by:
\begin{equation}
(G^{p}_t, G^{i}_t)(e) = \begin{cases}
    (1,1) & \text{w.p. } p_{11} \ \text{(shared edge)} \\
    (1,0) & \text{w.p. } p_{10} \ \text{(disagreement)} \\
    (0,1) & \text{w.p. } p_{01} \ \text{(disagreement)} \\
    (0,0) & \text{w.p. } p_{00} \ \text{(shared non-edge)}
\end{cases}
\end{equation}
where \(p_{11}+p_{10}+p_{01}+p_{00}=1\). For each object–viewpoint pair \(e \in \mathcal{E}_{\mathrm{a}}\), a similar distribution is defined with probabilities \(\boldsymbol{Q} = (q_{11}, q_{10}; q_{01}, q_{00})\). Given the perceived graph \(G^{p}_t\) and the imagined graph \(G^{i}_t\), the $\boldsymbol{P}, \boldsymbol{Q}$ can be approximated by Algorithm \ref{alg:pq}. Then we define:
\begin{align}
& \psi_{\mathrm{u}} \triangleq\left(\sqrt{p_{11} p_{00}}-\sqrt{p_{10} p_{01}}\right)^2, \\
& \psi_{\mathrm{a}} \triangleq\left(\sqrt{q_{11} q_{00}}-\sqrt{q_{10} q_{01}}\right)^2.
\end{align}

To make the navigator effective, we need to measure the confidence in the alignment between the perceived graph \(G^{p}_t\) and the imagined graph \(G^{i}_t\). 
We define:
\begin{itemize}
    \item An alignment is a mapping, or permutation, between the \(n\) objects in \(G^{p}_t\) and the \(m\) objects in \(G^{i}_t\). We denote a specific alignment by \(\pi\).
    \item \(\mathcal{S}_n\) is the set of all \(n!\) possible alignments.
    \item \(\pi_{\mathrm{id}}\) is the identity alignment, which represents the true, correct alignment where every object in \(G^{p}_t\) is matched to its corresponding object in \(G^{i}_t\).
    \item \(\delta_\pi(G^{p}_t, G^{i}_t)\) is the alignment difference. This score measures how much more likely the true alignment \(\pi_{\mathrm{id}}\) is compared to an incorrect alignment \(\pi\). If \(\delta_\pi > 0\), the previous alignment is a better fit.
\end{itemize}

A matching error occurs if there exists any incorrect alignment \(\pi \in \mathcal{S}_n \setminus \{\pi_{\mathrm{id}}\}\) that looks as good as, or better than, the true one. This happens when \(\delta_\pi(G^{p}_t, G^{i}_t) \leq 0\).

Therefore the probability of ambiguous alignment is defined as
\begin{equation}
  \mathrm{P}(\text{A}) = \mathrm{P}\left(\exists \pi \in \mathcal{S}_n \setminus\{\pi_{\mathrm{id}}\}, \delta_\pi\left(G^{p}_t, G^{i}_t\right) \leq 0\right).
\end{equation}

Calculating this probability exactly is often intractable. Instead, we rely on an upper bound from the latest graph alignment theory~\cite{AGA}, which limits the maximum possible value of this error probability. Given $\left(G^{p}_t, G^{i}_t\right)$ with n and m nodes, the upper bound can be computed by
\begin{equation}
    \mathrm{P}(\text{A}) \leq  e^{-2 \log n+2 m \psi_{\mathrm{a}}+2 n p_{11}}.
    \label{eq:error_bound}
\end{equation}
With this upper bound on the error, the navigator confidence \(C_t\) is given by
\begin{equation}
    C_t = 1 - \mathrm{P}(\text{A}) \geq 1 - e^{-2 \log n+2 m \psi_{\mathrm{a}}+2 n p_{11}}.
    \label{eq:confidence}
\end{equation}
If \(C_t > \tau_C\), the alignment between \(G^{p}_t\) and \(G^{i}_t\) is considered reliable with probability $C_t$, and the agent proceeds without invoking the costly slow system \(\Pi_{\mathrm{slow}}\), thereby achieving superior efficiency. The operation process of our SFCo-Nav is demonstrated in Algorithm \ref{alg:sfco-nav}.

\begin{algorithm}[t]
\caption{SFCo-Nav: Slow–Fast Collaborative Zero-Shot VLN}
\label{alg:sfco-nav}
\KwIn{Navigation instruction $I$}
\KwOut{Action sequence $\mathbf{A} = (a_0, a_1, \dots, a_T)$}
\BlankLine
\textbf{Initialize:} $G^{p}_0 \gets \emptyset$, $H_0 \gets \emptyset$;\\
Call $\Pi_{\mathrm{slow}}$ to identify final target $o_{\mathrm{goal}}$ via Eq.~(\ref{eq:f_goal});\\
Generate initial plan $\mathcal{P}_0 \gets f_{\mathrm{chain}}(f_{\mathrm{policy}}(I, H_0, G^{p}_0))$;\\

\For{$t \gets 1$ \textbf{to} $T$}{
    \textbf{1. Perception \& Local Graph Update:}\\
    \quad Build perceived object graph $G^{p}_t$ from onboard sensors;\\
    
    \textbf{2. Confidence Evaluation:}\\
    \quad Retrieve current imagined graph $G^{i}_t$ from $\mathcal{P}_{t-1}$;\\
    \quad Compute $C_t$ via Eq.~(\ref{eq:confidence}) for $G^{p}_t$ vs. $G^{i}_t$;\\
    
    \eIf{$C_t \leq \tau_C$}{
        \textbf{3a. Slow LLM Trigger:}\\
        \quad Append reached subgoals to $H_{t-1}$;\\
        \quad $(R_t, (o_t, sk_t)) \gets f_{\mathrm{policy}}(I, H_{t-1}, G^{p}_t)$;\\
        \quad $\mathcal{P}_t \gets f_{\mathrm{chain}}(R_t)$;\\
    }{
        \textbf{3b. Fast Navigator Execution:}\\
        \quad Keep $\mathcal{P}_{t-1}$ without triggering LLM;\\
    }
    
    \textbf{4. Reactive Navigation:}\\
    \quad $a_t \gets \pi_{\mathrm{fast}}(G^{p}_t, o_t, sk_t)$; Execute $a_t$ \\ 
    \quad Pop the first executed item of $\mathcal{P}_{t-1}$;\\
    
    \If{Reached $o_{\mathrm{goal}}$ within $d_{\mathrm{stop}}$}{
        \textbf{Terminate} and return $\{a_k\}_{k=0}^t$; \textbf{break};
    }
}
\end{algorithm}





\begin{table*}[]
    \centering
    \caption{Comparison on the R2R Dataset.}
    \renewcommand\arraystretch{1.2}
    \belowrulesep=0pt
    \aboverulesep=0pt
\resizebox{0.8\textwidth}{!}{
\begin{tabular}{c|ccccc|ccc|ccc}
\toprule
\multirow{2}{*}{Method} & \multicolumn{5}{c|}{Task Success Metric} & \multicolumn{3}{c|}{Token Efficiency Metric} & \multicolumn{3}{c}{Time Efficiency Metric[s]}  \\
\cmidrule(lr){2-6} \cmidrule(lr){7-9} \cmidrule(lr){10-12} 
& TL & NE & OSR & SR & SPL & V-Tok & L-Tok & U-Tok & V-Time & L-Time & T-Time \\

\cmidrule{1-12}
NavGPT (GPT-4) & 11.45 & 6.46 & 42.0 & 34.0 & 29.0 & 108k & 64.62k & 496.62k & 1440 & 48.19 & 1488.19 \\
MapGPT (GPT-4) & -- & 6.29 & \textbf{57.6} & 38.8 & 25.8 & 4.95k & \uline{9.46k} & 29.26k & 66 & 41.01 & 107.01 \\
NavCoT (Tuned$^*$ Llama) & \textbf{9.83} & 6.67 & 44.0 & 36.4 & 33.17 & 4.50k & \textbf{1.42k} & 19.42k & 60 & \textbf{16.42} & 76.42 \\

\cmidrule{1-12}
\textbf{SF-Nav (GPT-4o)} & 10.89 & \textbf{5.70} & \uline{53.8} & \textbf{41.3} & \textbf{36.35} & \textbf{0.9k} & 13.10k & \uline{16.10k} & \textbf{3.36} & 36.0 & \uline{39.36} \\
\textbf{SFCo-Nav (GPT-4o)} & \uline{10.80} & \uline{6.04} & 50.5 & \uline{38.2} & \uline{32.54} & \textbf{0.9k} & 9.94k & \textbf{13.54k} & \textbf{3.36} & \uline{22.8} & \textbf{26.16} \\

\bottomrule
\end{tabular}}
\label{tab:MTR_TTR}
\end{table*}

\begin{table*}[]
    \centering
    \caption{Comparison on the REVERIE Dataset.}
    \renewcommand\arraystretch{1.2}
    \belowrulesep=0pt
    \aboverulesep=0pt
\resizebox{0.8\textwidth}{!}{
\begin{tabular}{c|ccccc|ccc|ccc}
\toprule
\multirow{2}{*}{Method} & \multicolumn{5}{c|}{Task Success Metric} & \multicolumn{3}{c|}{Token Efficiency Metric} & \multicolumn{3}{c}{Time Efficiency Metric[s]}  \\
\cmidrule(lr){2-6} \cmidrule(lr){7-9} \cmidrule(lr){10-12} 
& TL & NE & OSR & SR & SPL & V-Tok & L-Tok & U-Tok & V-Time & L-Time & T-Time \\
\cmidrule{1-12}
NavGPT (GPT-4) & -- & -- & 28.3 & 19.2 & 14.6 & 108.00k & 57.35k & 489.35k & 1440 & 49.74 & 1489.74 \\
MapGPT (GPT-4) & -- & -- & \uline{42.6} & 28.4 & 14.5 & 4.95k & \uline{7.56k} & 27.36k & 66 & 26.57 & 92.57 \\
NavCoT (Tuned$^*$ Llama) & 12.36 & -- & 14.20 & 9.20 & 7.18 & 4.50k & \textbf{1.44k} & 19.44k & 60 & \textbf{11.40} & 71.40 \\
\cmidrule{1-12}
\textbf{SF-Nav (GPT-4o)} & \textbf{10.05} & \uline{7.87} & \textbf{46.63} & \textbf{35.87} & \textbf{30.62} & \textbf{0.9k} & 14.04k & \uline{17.64k} & \textbf{3.36} & 31.4 & \uline{34.76} \\
\textbf{SFCo-Nav (GPT-4o)} & \uline{10.32} & \textbf{7.16} & 42.36 & \uline{31.33} & \uline{27.01} & \textbf{0.9k} & 9.44k & \textbf{13.04k} & \textbf{3.36} & \uline{21.6} & \textbf{24.96} \\
\bottomrule
\end{tabular}}
\label{tab:MTR_TTR_Reverie}
\end{table*}




\section{EXPERIMENTS}
In this section, SFCo-Nav is evaluated against some state-of-the-art zero-shot VLN systems. First, we introduce the experimental setup, evaluation metrics, and VLN environments used for testing. Then, experiments conducted on open datasets is presented, focusing on two key aspects: 1) the overall task success performance, and 2) the token/temporal computational efficiency of SFCo-Nav.
\subsection{Experimental Setup}
\subsubsection{Evaluation Metrics}
The VLN systems are evaluated on three levels: task effectiveness, token efficiency, and temporal efficiency.

\textit{Task-Level Metrics} evaluate navigation effectiveness and path efficiency using standard VLN metrics\cite{R2R}. Success Rate (SR) measures task completion. Success rate weighted by Path Length (SPL) measures success and efficiency. Navigation Error (NE) measures final distance to goal. Oracle Success Rate (OSR) measures trajectory quality. Trajectory Length (TL) measures path length.

\textit{Token-Level Metrics} quantify the computational cost to validate efficiency gains.
\begin{itemize}
    \item Language Tokens per Path (L-Tok): Average number of LLM tokens processed per episode, measuring slow-system cost.
    \item Vision Tokens per Path\footnote{Visual token counts are measured based on GPT-4o tokenizer.} (V-Tok): Average number of vision tokens processed per episode, measuring visual perception load.
    \item Unified Tokens per Path (U-Tok): A combined metric representing overall token throughput, calculated as
    \begin{equation}
        \text{U-Tok} = \text{L-Tok} + \lambda \times \text{V-Tok},
    \end{equation}
    where $\lambda$ is a cost coefficient that weights vision tokens relative to language tokens. 
    Following OpenAI’s reported inference costs\footnote{The token price of OpenAI may refer to \url{https://platform.openai.com/docs/pricing##image-tokens}.}, we set $\lambda = 4$ to reflect that each vision token is approximately four times more expensive to process than a text token.
\end{itemize}

\textit{Timing-Level Metrics} measure wall-clock latency for real-world deployment.
\begin{itemize}
    \item LLM Time per Path (L-Time): Total wall-clock time (s) for all LLM inference calls per episode, capturing slow-system latency.
    \item Vision Time per Path\footnote{Visual perception time for our method is measured using Grounding‑DINOv2\cite{GroundingDINO}, whereas VLM‑based baselines use BLIP‑2.} (V-Time): Total wall-clock time (s) for all visual processing per episode, capturing visual perception latency.
    \item Total Time per Path (T-Time): A unified metric for the overall episode latency, calculated as
\begin{equation}
\text{T-Time} = \text{L-Time} + \text{V-Time}.
\end{equation}
\end{itemize}

\subsubsection{Development of Experimental Datasets}
To demonstrate the effectiveness of SFCo-Nav, experiments are conducted on public visual language navigation datasets and real-world suite environment.

\textit{R2R}\cite{R2R}: The standard benchmark for VLN, requiring agents to follow detailed path-based instructions in photorealistic indoor environments.

\textit{REVERIE}\cite{Reverie}: A more challenging object-grounding benchmark where agents follow high-level, goal-oriented instructions to find a specific target object.

\textit{Real-world Suite}: Deployment on a physical legged robot in an indoor hotel suite setting to test the system's robustness and practical performance under real-world conditions.

\subsubsection{Compared Algorithms}
SFCo-Nav is compared against latest (partial) zero-shot VLM-LLM based visual language navigation methods as follows:

\begin{itemize} 
\item NavGPT\cite{NavGPT}: A foundational approach that uses an LLM as a zero-shot planner, making decisions based entirely on textual descriptions of visual observations obtained by VLM.
\item MapGPT\cite{MapGPT}: This method augments the LLM planner with an explicit spatial memory by constructing a real-time topological map and feeding it to the LLM in a textual format obtained by VLM for navigation.
\item NavCoT\cite{navcot}: This approach utilizes Chain-of-Thought (CoT) prompting to improve reasoning by instructing the LLM to generate an explicit thought process before selecting an action, which is based on zero-shot VLM and fine-tuned LLM.
\item SF-Nav: Our proposed slow–fast framework variant without the slow–fast bridge for asynchronous LLM triggering. In this setting, the slow LLM planner is invoked at every navigation step. 
\end{itemize}

\subsection{Experimental Results and Discussions}



\subsubsection{R2R Dataset}
On the R2R benchmark, SFCo-Nav achieves a strong balance between accuracy and efficiency, as shown in Table \ref{tab:MTR_TTR}. It achieves competitive task success metrics that closely match or surpass prior zero-shot VLN baselines, while drastically reducing computational cost. Compared to NavGPT, MapGPT, NavCoT, SF-Nav, SFCo-Nav achieves the lowest unified token usage, thanks to its asynchronous slow–fast triggering mechanism. This reduction directly translates into the fastest total inference time, more than 4× faster than MapGPT and 33\% faster than SF-Nav, without sacrificing navigation quality. These results confirm that SFCo-Nav delivers near state-of-the-art accuracy with unmatched token and time efficiency.
\subsubsection{REVERIE Dataset}
On the REVERIE dataset, SFCo-Nav achieves an NE of 7.16 meters and SR of 31.33\%, ranking second only to SF-Nav,  while outperforming all other baselines, as shown in Table \ref{tab:MTR_TTR_Reverie}. In terms of efficiency, SFCo-Nav reduces LLM token consumption to 9.44k, a 33\% drop compared with SF-Nav, yielding the lowest unified token usage among all methods. This compact token footprint translates into the fastest total inference time at 24.96 seconds, over 3.7× faster than MapGPT and more than 59× faster than NavGPT. These results confirm that SFCo-Nav retains competitive task performance while achieving substantial gains in computational and time efficiency.  
\begin{figure*}
    \centering
    \includegraphics[width=0.9\linewidth]{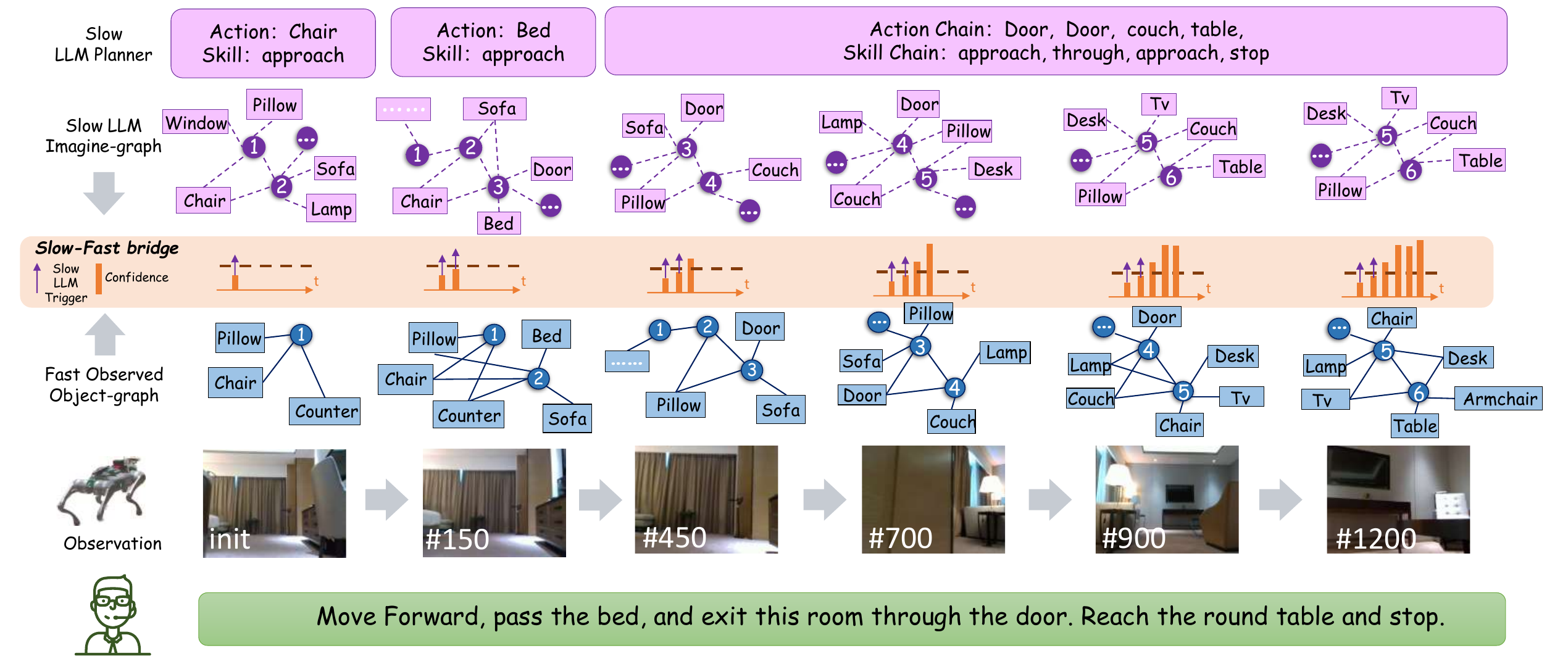}
    \caption{Real-world hotel suite deployment of SFCo-Nav. Early in navigation, sparse observations yield low match probability, triggering the slow planner. As observations grow, confidence exceeds the threshold, enabling fast, LLM-free execution. This slow-fast collaboration preserves success while improving time efficiency and reducing token usage.}
    \label{fig:case_study}
    \vspace{-0.5cm}
\end{figure*}
\subsubsection{Ablation Study}
\begin{table}[]
    \centering
    \caption{Comparison of Various Confidence level on  R2R Subset.}
    \renewcommand\arraystretch{1.2}
    \belowrulesep=0pt
    \aboverulesep=0pt
\resizebox{0.45 \textwidth}{!}{
\begin{tabular}{c|ccc|cc}
\toprule
\multirow{2}{*}{\makecell{SFCo-Nav-\\Confidence Threshold}}  & \multicolumn{3}{c|}{Task Success Metric} & \multicolumn{2}{c}{Efficiency Metric}  \\
\cmidrule(lr){2-4} \cmidrule(lr){5-6} & SR & OSR & SPL & U-Tok & T-Time[s]  \\

\cmidrule{1-6}
SFCo-Nav-1.0 & \textbf{39.67} & 49 & \textbf{37.08} &12.58k &28.08\\
SFCo-Nav-0.95 & 36.67 & \textbf{50.33} & 34.10 &10.55k & 22.5\\
SFCo-Nav-0.85 & 36.67 & \textbf{50.33} & 34.03 &10.42k & 22.2 \\
SFCo-Nav-0.6 & 33.33 & 48.33 & 31.22 &9.77k & 20.34\\
SFCo-Nav-0.4 & 29.67 & 45 & 27.39  & \textbf{9.51k}& \textbf{19.64}\\

\bottomrule
\end{tabular}}
\label{tab:MTR TTR}
\vspace{-0.7cm}
\end{table}

\begin{table}[]
    \centering
    \caption{Ablation Study of SFCo-Nav on R2R Subsets.}
    \renewcommand\arraystretch{1.2}
    \belowrulesep=0pt
    \aboverulesep=0pt
\resizebox{0.9\linewidth}{!}{
\begin{tabular}{c|c|c|c|c|c}
\toprule
LLMs & \makecell{Chain \\ Decision} & \makecell{Object-Skill \\ Pair} & SR & U-Tok & T-Time[s] \\
\cmidrule{1-6} \multirow{4}{*}{GPT-4o}
  & $\times$ & $\times$ & 32 & 13.29k & 32.90 \\
  & \checkmark & $\times$ & 31.33 & 10.13k & 28.73 \\
 & $\times$ & \checkmark & 45.67 & 13.06k & 31.05\\
 & \checkmark & \checkmark  & 37 & 9.9k & 27.47 \\
 \cmidrule{1-6} \multirow{4}{*}{Deepseek-V3.1}
  & $\times$ & $\times$ & 32.67 & 15.07k &  99 \\
  & \checkmark & $\times$ & 31 & 12.87k &  96.84\\
 & $\times$ & \checkmark & 41 & 15.91k & 102.24\\
 & \checkmark & \checkmark  & 34 & 12.52k & 97.56 \\

\bottomrule
\end{tabular}}
\label{tab:ablation}
\vspace{-0.5cm}
\end{table}
Ablation study is conducted on two main aspects: a) Impact of the confidence threshold. b) Imagined Chain and Object-Skill pair module ablation with different pre-trained LLM. To reduce evaluation cost, a subset of R2R including 300 instruction trajectories is used for this ablation study.
\paragraph{Impact of the confidence threshold}
We investigate the impact of the confidence threshold on SFCo-Nav’s performance by varying the trigger level for invoking the slow LLM planner from 1.0 to 0.4. As shown in Table \ref{tab:MTR TTR}, higher thresholds generally yield stronger task success metrics but incur higher token and time costs. Specifically, SFCo-Nav-1.0 achieves the highest SR and SPL but requires 12.58k total tokens and 28.08 seconds per trajectory. Conversely, SFCo-Nav-0.4 minimizes computation, with the smallest total tokens and fastest runtime, but SR drops to 29.67\% and SPL to 27.39\%. Intermediate thresholds 0.85–0.95 offer a better balance, reducing runtime by 20\% while maintaining SR within 3\% of the maximum. These results highlight a clear accuracy–efficiency trade-off governed by the confidence threshold.
\paragraph{Ablation on Imagined Chain and Object–Skill Pair Modules with different LLM}
We assess the impact of the Imagined Chain and Object–Skill Pair modules in SFCo-Nav using GPT‑4o and Deepseek‑V3.1 backbones, as shown in Table \ref{tab:ablation}. Across both LLMs, the Object–Skill Pair consistently boosts SR, while the Imagined Chain mainly reduces token usage and runtime with minor SR changes. Combining both yields a balanced trade-off, for GPT‑4o with the lowest token usage and fastest runtime; for Deepseek, tokens drop from 15.07k to 12.52k with only a slight SR decrease. These results indicate that the Object–Skill Pair improves navigation quality, while the Imagined Chain enhances efficiency.

\subsubsection{Real-world Suite Case Study}

To assess real‑world deployability, we implemented SFCo‑Nav on a legged robot navigating a furnished hotel suite, as depicted in Fig.~\ref{fig:case_study}. In early steps, few observed objects yield low match probability and confidence, triggering the slow LLM planner for planning and multi‑step imagination. As observations grow, confidence surpasses the threshold. This enables the fast module to execute without additional LLM calls. This asynchronous slow–fast triggering mechanism preserves navigation success while improving time efficiency and reducing token use. It demonstrates SFCo‑Nav’s capability for efficient and accurate real‑world navigation.

\section{CONCLUSIONS}
In this work, we introduce SFCo‑Nav, a slow–fast collaborative framework for zero‑shot visual language navigation that couples a slow LLM‑based planner with a fast reactive navigator via an asynchronous confidence‑triggered bridge. Experiments on R2R and REVERIE show that SFCo‑Nav achieves competitive or superior success rates with significantly reduced inference cost. A real‑world deployment on a legged robot in a furnished hotel suite demonstrates its ability to adaptively invoke LLM reasoning only when necessary, maintaining navigation accuracy while improving time efficiency and lowering token usage. These results highlight SFCo‑Nav’s potential for practical, efficient embodied AI in real‑world environments.

\addtolength{\textheight}{-12cm}   





\normalem
\bibliographystyle{IEEEtran}
\bibliography{ref}

\end{document}